\pdfoutput=1
\documentclass[letterpaper, conference]{IEEEtran}
\IEEEoverridecommandlockouts

\usepackage[letterpaper, top=0.75in, bottom=1in, left=0.625in, right=0.625in]{geometry}
\usepackage{microtype}

\usepackage{cite}
\usepackage{amsmath,amssymb,amsfonts}
\usepackage{graphicx}
\usepackage{textcomp}
\usepackage{xcolor}
\usepackage{booktabs}
\usepackage{url}
\usepackage{placeins}
\usepackage{balance}
\usepackage{tikz}
\usetikzlibrary{arrows.meta,shapes.geometric,positioning,fit,backgrounds,calc}

\definecolor{llmblue}   {RGB}{76,114,176}
\definecolor{vitgreen}  {RGB}{85,168,104}
\definecolor{vlmred}    {RGB}{196,78,82}
\definecolor{fedorange} {RGB}{204,185,116}
\definecolor{fedpurple} {RGB}{129,114,178}
\definecolor{outyyellow}{RGB}{255,242,204}
\definecolor{smallgray} {RGB}{220,220,220}
\definecolor{fignavy}   {RGB}{1,41,84}

\graphicspath{{generated/}}

\tikzset{
  mainbox/.style={draw, rounded corners=4pt, align=center,
                  minimum height=0.75cm, text width=#1,
                  font=\footnotesize, inner sep=4pt},
  mainbox/.default=2.4cm,
  llmbox/.style ={mainbox, fill=llmblue!30,  draw=llmblue!80!black},
  vitbox/.style ={mainbox, fill=vitgreen!30, draw=vitgreen!80!black},
  vlmbox/.style ={mainbox=5.2cm, fill=vlmred!25, draw=vlmred!80!black,
                  minimum height=1.0cm},
  fedbox/.style ={mainbox=1.9cm, fill=fedpurple!40, draw=fedpurple!80!black},
  outbox/.style ={mainbox=4.5cm, fill=outyyellow, draw=yellow!70!black},
  sbox/.style   ={draw, rounded corners=2pt, fill=smallgray,
                  minimum height=0.5cm, align=center,
                  font=\scriptsize, inner sep=3pt, text width=1.8cm},
  arr/.style    ={-{Stealth[length=5pt]}, thick},
  darr/.style   ={-{Stealth[length=5pt]}, thick, dashed},
}

\begin{document}
\renewcommand{\arraystretch}{0.85}
\setlength{\abovecaptionskip}{0pt}
\setlength{\belowcaptionskip}{0pt}

\setlength{\textfloatsep}{6pt plus 1pt minus 1pt}
\setlength{\dbltextfloatsep}{6pt plus 1pt minus 1pt}
\setlength{\intextsep}{8pt plus 2pt minus 2pt}
\setlength{\floatsep}{4pt plus 1pt minus 1pt}
\setlength{\dblfloatsep}{4pt plus 1pt minus 1pt}

\setlength{\columnsep}{0.25in}

\renewcommand{\dbltopfraction}{0.95}
\renewcommand{\topfraction}{0.95}
\renewcommand{\textfraction}{0.05}
\renewcommand{\floatpagefraction}{0.75}
\renewcommand{\dblfloatpagefraction}{0.75}

\title{OmniMed-FL: A Robust Multimodal Federated Learning Framework for Clinical Diagnosis}

\author{
\IEEEauthorblockN{$^1$Ayush Debnath, $^{3,4}$Ruelia Saha, $^{2}$Sudip Misra}
\IEEEauthorblockA{$^{1,2}$Indian Institute of Technology Kharagpur, India\enspace $^3$KTH Royal Institute of Technology, Sweden\enspace $^4$SRM University-AP, India\\
Emails: \{ayush.d@kgpian.iitkgp.ac.in, ruelia.saha.rs@gmail.com,
sudipm@iitkgp.ac.in\}}
}

\maketitle

\begin{abstract}
Simultaneous assessment of medical imaging and
patient records is often required in clinical diagnosis. However, standard
machine learning algorithms cannot analyze these several sorts
of data together. Meanwhile, compliance with the Health Insurance Portability and
Accountability Act (HIPAA) and the General Data Protection Regulation (GDPR) can
constrain centralized aggregation of sensitive patient data. This leaves
a crucial void of secure fusion of visual and textual context
across distant networks. Thus, we present OmniMed-FL,
a controlled systems study of multimodal federated learning for five-class clinical
condition classification (Normal, Pneumonia, COVID-19, Pleural Effusion,
Cardiomegaly). Our proxy corpus pairs 3,000 public chest
radiographs with 3,000 class-conditioned synthetic notes, matched by
class, not by patient. The framework benchmarks eight fusion strategies,
three initializations, four missing-text imputation rules, and matched federated
baselines under non-IID Dirichlet partitioning across 3 to 20 hospital clients.
As all notes are synthetic and pairing is not
patient-level, these are descriptive proxy comparisons, not estimates of diagnostic
performance or deployment readiness. Within those limits with clients($K{=}5$) and severe skew
($\alpha{=}0.1$), local-only training achieves a macro-F1 score of 0.297, FedAvg
achieves $0.662\!\pm\!0.074$, FedProx $0.737\!\pm\!0.085$, a matched FedMME-style
one-shot ensemble $0.647\!\pm\!0.080$, and our SCAFFOLD--AdamW adaptation
$0.070\!\pm\!0.015$, the 0.075 FedProx--FedAvg gap falling inside the wider of the
two two-seed standard deviations. Over a $4\!\times\!3$ grid, label skew costs up to 0.27 F1
whereas a near-sevenfold client increase costs at most 0.10, while bidirectional volume grows
linearly to 183.5\,GiB at $K{=}20$. Multimodal fusion leads on both corpora, scoring
0.956 against 0.934 for text and 0.664 for images on the synthetic corpus and 0.906
against 0.880 and 0.737 on the radiograph corpus, for $2.3\times$ the model state of
text alone.
\end{abstract}

\begin{IEEEkeywords}
Multimodal Federated Learning, Vision--Language Models, Clinical Condition
Classification, E-Health, Non-Identically Distributed Data, Communication Cost,
Retrieval-Augmented Generation
\end{IEEEkeywords}

\section{Introduction}

\IEEEPARstart{C}{linical} condition classification, which covers distinguishing
normal presentations from acute infections, fluid accumulation, and structural
cardiac pathology, is central to timely patient intervention. Automated systems
typically operate on a single modality. For example, vision models trained on chest
radiographs~\cite{Irvin2019} do not capture the full diagnostic picture. Image-only models miss
contextual cues such as BNP elevation or viral contact history, while text-only
models miss radiological patterns such as bilateral ground-glass opacity or pleural
fluid meniscus.

The healthcare setting complicates deployment further because patient data is
distributed across hospitals, clinics, and diagnostic centers.
Federated learning (FL) addresses these constraints by
transmitting only model updates~\cite{McMahan2017}, although keeping raw records at
home is a data-locality property, not a privacy proof. The cited clinical FL
systems are single-modality~\cite{Sheller2020,Dou2021},
discarding the notes clinicians record alongside imaging. The aforementioned lacunae is addressed by OmniMed-FL, so hospitals can collaborate on multimodal clinical models without
exposing patient data. We measure four properties of such a system, namely how far
label skew degrades accuracy, the point at which a five-class head collapses onto its
majority class, which fusion operator earns the compute it consumes, and what a
multimodal model costs to move. Each is measured on a controlled proxy benchmark
whose notes are synthetic and paired by class, so what we characterize is the
learning system rather than clinical diagnosis. Fusion scores
highest on both corpora, 0.956 on synthetic images and 0.906 on radiographs against
0.934 and 0.880 for text alone.

\subsection{Contribution}
The main contributions of this paper are as follows:
\begin{itemize}
\item We demonstrate that fusing radiographs with clinical notes improves accuracy across decentralized clients. Through extensive simulation under matched data, optimizer, and local budget, our multimodal approach significantly
outperforms isolated, single-modality systems.
\item We build a controlled five-class corpus whose notes mix wording across classes, so the label cannot be read off a single keyword, and we test how much of the label a model can still recover from wording alone.
\item We quantify the cost of keeping data locally. Federated training retains 94\% of a pooled control at moderate skew, and we show that severe label skew, not federation itself, is what erodes accuracy, alongside the runtime, memory, and communication each configuration spends.
\end{itemize}

\section{Related Work}

\subsection{Medical Image Classification}
Early automated diagnostics treated the chest radiograph as a pure computer-vision
problem. CheXNet and CheXpert set centralized CNN
baselines~\cite{Rajpurkar2017chexnet,Irvin2019}, and Vision Transformers later
established an attention-based alternative for general image
classification~\cite{Dosovitskiy2021}. All of them
pool data on one server, and none reads the note a clinician wrote next to the image.

\subsection{Federated Clinical Learning}
Privacy regulation moved the clinical community toward FL. FedAvg established
decentralized training~\cite{McMahan2017}, and later work added multi-institutional
collaboration without shared patient records~\cite{Sheller2020}, multinational
COVID-19 CT validation~\cite{Dou2021}, federated weakly supervised medical-image
segmentation~\cite{Lin2025fedlppa}, and communication-efficient health
monitoring~\cite{Chu2022globecom}. FedProx and SCAFFOLD address the client heterogeneity
  that degrades naive averaging~\cite{Li2020fedprox,Karimireddy2020,Tian2022globecom}. The clinical
  deployments discussed above remain unimodal; therefore, their behavior after
  incorporating a second modality remains unexplored.

\subsection{Vision--Language Models in Healthcare}
Centralized vision--language models advanced in parallel. BiomedCLIP, LLaVA-Med, and
MedCLIP jointly model biomedical visual and textual information using paired,
instruction-derived, or deliberately unpaired corpora,
respectively~\cite{Zhang2025biomedclip,Li2023llavamed,Wang2022medclip}, but they
train on centrally assembled corpora. Multimodal federated learning is
more recent. FedMME combines one-shot client models by voting, and P-FIN fills in
missing features with calibrated
uncertainty~\cite{Wang2025fedmme,Shahid2026pfin}, each on its own data and budget,
while cross-modal federated medical imaging is also
appearing~\cite{Yan2024crossmodal}.
Neither line reports how much bandwidth and memory a multimodal federated model
costs. 

\textit{Synthesis.} The landscape above offers either the multimodal accuracy of
centralized vision--language models or the data locality of federated learning, but
not both. OmniMed-FL bridges that gap, embedding text--image fusion inside a
federated loop that leaves every record at the hospital that holds it.

\section{System Design}

\subsection{Architecture}

\begin{figure*}[t]
\centering
\begin{minipage}[t]{0.40\textwidth}
\centering
\begin{tikzpicture}
  \node[inner sep=0pt, anchor=south west] (pa)
        {\includegraphics[width=\linewidth]
                         {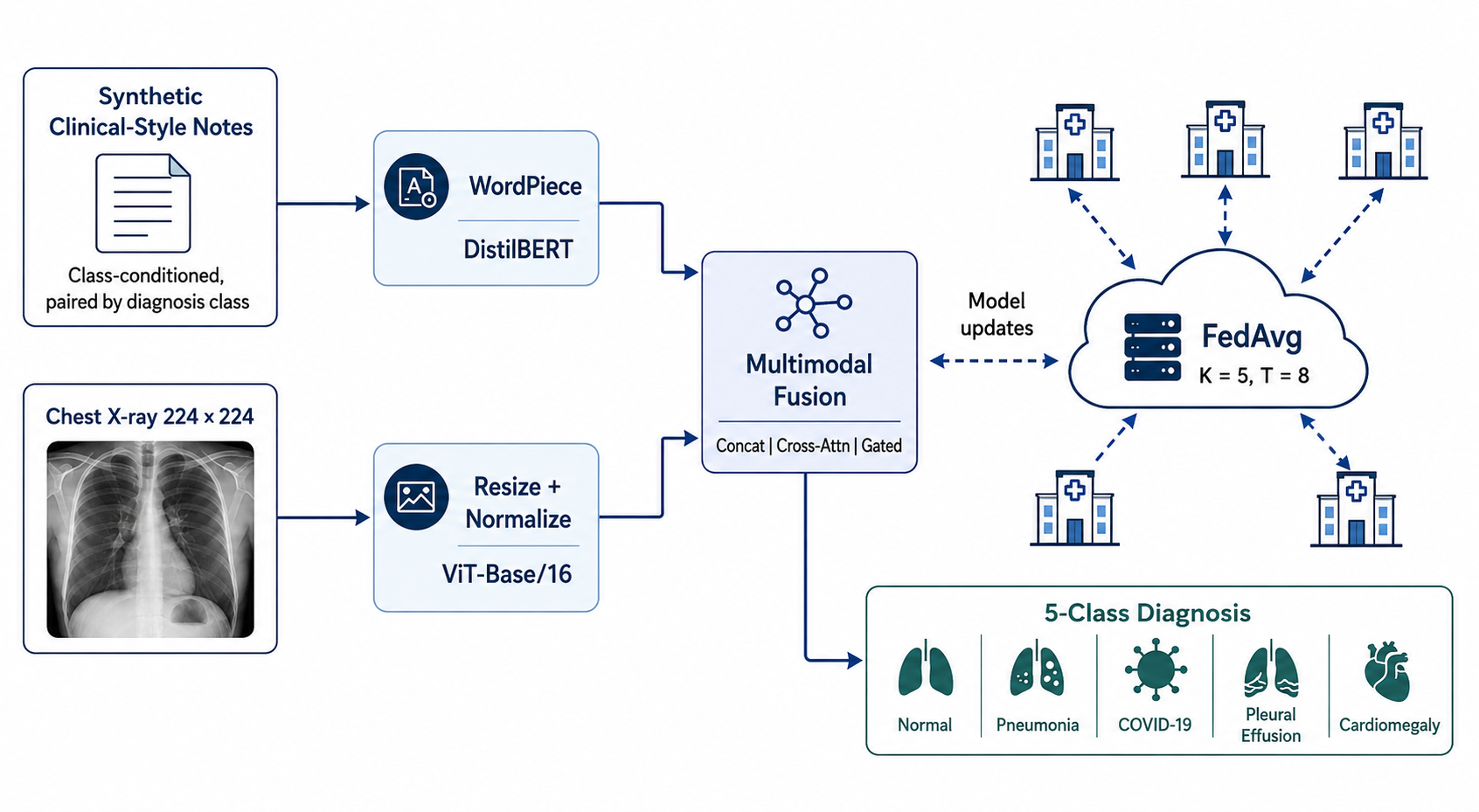}};
  \begin{scope}[x={(pa.south east)}, y={(pa.north west)}]
    \draw[dashed, semithick, rounded corners=2pt, draw=fignavy]
      (0.008,0.968) -- (0.625,0.968) -- (0.625,0.303) -- (0.992,0.303) --
      (0.992,0.045) -- (0.008,0.045) -- cycle;
    \coordinate (pabadge) at (0.075,0.968);
    \node[inner sep=0pt, text=fignavy,
          font=\sffamily\bfseries\fontsize{5.5}{6}\selectfont]
          at (0.822,0.653) {Server};
  \end{scope}
  \node[anchor=west, fill=white, inner xsep=2pt, inner ysep=0.5pt,
        font=\scriptsize\bfseries, text=fignavy]
        at (pabadge) {Hospital Client $k$ (local computation)};
  \node[fill=white, inner sep=1pt, anchor=east] at (pabadge)
        {\includegraphics[height=0.36cm]{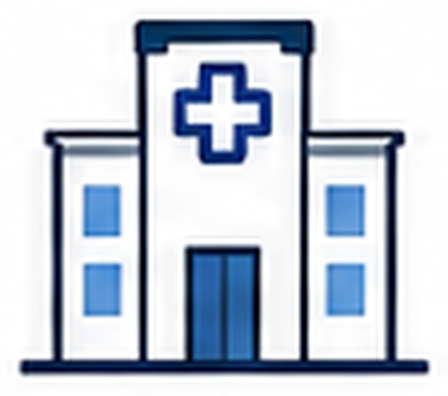}};
\end{tikzpicture}
\par\vspace{1pt}\textbf{(a)}
\end{minipage}\hfill
\begin{minipage}[t]{0.40\textwidth}
\centering
\resizebox{0.98\linewidth}{!}{%
\begin{tikzpicture}[node distance=0.45cm]
  \node[llmbox] (ti)  {Text Input\\[-1pt]{\scriptsize Template clinical-style note}};
  \node[sbox, below=0.28cm of ti] (tok) {Tokenizer\\WordPiece};
  \node[llmbox, below=0.28cm of tok, text width=2.8cm, minimum height=1.0cm]
        (llm) {Text Encoder\\[-2pt]{\scriptsize DistilBERT}};
  \node[font=\scriptsize, below=0.18cm of llm] (ht)
        {$\mathbf{h}_t\!\in\!\mathbb{R}^{256}$};

  \node[vitbox, right=3.4cm of ti] (ii) {Image Input\\[-1pt]
                                         {\scriptsize Chest X-ray $224{\times}224$}};
  \node[sbox, below=0.28cm of ii] (aug) {Resize\\Normalize};
  \node[vitbox, below=0.28cm of aug, text width=2.8cm, minimum height=1.0cm]
        (vit) {Image Encoder\\[-2pt]{\scriptsize ViT-Base/16}};
  \node[font=\scriptsize, below=0.18cm of vit] (hv)
        {$\mathbf{h}_v\!\in\!\mathbb{R}^{512}$};

  \node[fedbox, right=2.1cm of vit, yshift=0.3cm, minimum height=1.4cm,
        text width=2.0cm] (fed)
        {Aggregation Server\\[-2pt]\textbf{FedAvg}\\[-1pt]
         {\scriptsize $K{=}3$--20 clients\\$T{=}8$ rounds}};

  \node[vlmbox, below=1.0cm of llm, xshift=3.1cm, minimum height=0.95cm]
        (vlm) {Multimodal Fusion\\[-2pt]
               {\scriptsize Concat $|$ Cross-Attn $|$ Gated $|$ Dual-Proj $|$
               Decoder A/B $|$ Cross-Attn-384 $|$ Token Encoder}};

  \node[mainbox=3.2cm, fill=vlmred!40, draw=vlmred,
        below=0.35cm of vlm] (cls)
        {Classification Head (5-class)\\[-1pt]
         {\scriptsize Focal + entropy/confidence penalties}};

  \node[outbox, below=0.35cm of cls] (out)
        {Normal $|$ Pneumonia $|$ COVID-19\\[-2pt]
         {\scriptsize Pleural Effusion $|$ Cardiomegaly}};

  \begin{scope}[on background layer]
    \node[draw=llmblue!80!black, fill=llmblue!4, rounded corners=6pt,
          dashed, thick, fit=(ti)(tok)(llm)(ht)(ii)(aug)(vit)(hv)(vlm)(cls)(out),
          inner sep=7pt] (client) {};
  \end{scope}
  \node[anchor=south west, fill=white, inner xsep=3pt, inner ysep=1pt,
        font=\scriptsize\bfseries, text=llmblue!80!black]
        (clientlbl) at ([xshift=1.05cm]client.north west)
        {Hospital Client $k$ (local computation)};
  \node[fill=white, inner sep=1pt, anchor=east] at ([xshift=-1pt]clientlbl.west)
        {\includegraphics[height=0.70cm]{generated/icon_hospital_client.png}};

  \draw[arr](ti)--(tok);\draw[arr](tok)--(llm);\draw[arr](ii)--(aug);
  \draw[arr](aug)--(vit);\draw[arr](llm)--(ht);\draw[arr](vit)--(hv);
  \draw[arr](ht.south)--++(0,-0.15)-|(vlm.north west);
  \draw[darr](hv.south)--++(0,-0.15)-|(vlm.north east);
  \draw[arr](vlm)--(cls);\draw[arr](cls)--(out);
  \draw[<->, >=Stealth, thick, dashed]
        (client.east |- fed.center)--node[above, font=\tiny, align=center,
              inner sep=1.5pt]{model\\updates}(fed.west);
\end{tikzpicture}%
}
\par\vspace{1pt}\textbf{(b)}
\end{minipage}
\caption{OmniMed-FL architecture. (a) Overall multimodal federated workflow at the
reference $K{=}5$ setting; the dashed boundary encloses what runs inside a single
hospital client, and only model updates cross it to the FedAvg cloud shared by the
other clients. (b) Detailed architecture with the same client-side computation
boxed separately from the FedAvg aggregation server.}
\label{fig:arch}
\end{figure*}

Figure~\ref{fig:arch} illustrates the pipeline of the proposed scheme. We tokenize notes with WordPiece at
a 128-token cap, resize radiographs to $224{\times}224$, and normalize them
against ImageNet statistics. The DistilBERT~\cite{Sanh2019} \texttt{[CLS]} vector
is projected to $\mathbf{h}_t\!\in\!\mathbb{R}^{256}$, and mean-pooled ViT tokens
give $\mathbf{h}_v\!\in\!\mathbb{R}^{512}$. Eight fusion rules then combine
the text and image vectors. Three use no attention (projected concatenation,
dual-projection concatenation, and gated residual fusion) and five do (residual
cross-attention, its non-residual 384-d variant, a one-layer decoder A, a
two-query/two-layer decoder B, and a two-token Transformer encoder). All eight feed
the same two-layer, five-class multilayer perceptron
(MLP) and share the data, partition, optimizer, and local budget. The eight rules test
whether the attention-free ones score as well as the attention-based ones when both
get the same training on hardware one hospital can afford.

\subsection{Federated Learning Setup}
\label{sec:objective}
Let $K$ clients hold disjoint shards $\mathcal{D}_k$ of the training split, with
$n_k=|\mathcal{D}_k|$ and $n=\sum_k n_k$, and let $\theta$ collect the trainable
parameters on the fusion path: both encoders, the projection layers, the fusion
operator, and the classification head. We minimize the sample-weighted empirical
risk
\begin{equation}
\begin{aligned}
\min_{\theta}\ F(\theta)&=\sum_{k=1}^{K}\frac{n_k}{n}F_k(\theta),\\[-2pt]
F_k(\theta)&=\frac{1}{n_k}\!\!\sum_{(x,z,y)\in\mathcal{D}_k}\!\!\ell\big(f_\theta(x,z),y\big).
\end{aligned}
\label{eq:fedobj}
\end{equation}
where $x$ is a radiograph, $z$ its class-paired note, $y$ the class label, and
$f_\theta$ the fused five-class classifier. Shards never leave a client, and the
server sees only the trainable tensors listed above. For mini-batch
$\mathcal{B}$, the loss we actually implement is
\begin{equation}
\ell_{\mathcal B}=\frac{1}{|\mathcal B|}\sum_{i\in\mathcal B}
(1-p_{i,y_i})^{\gamma}\mathrm{CE}_{\epsilon,i}
+\lambda\phi(h)+5[q-0.9]_+,
\label{eq:loss}
\end{equation}
with $p_{i,y_i}$ the true-class probability, $\mathrm{CE}_{\epsilon,i}$
label-smoothed cross-entropy, $\gamma{=}2$, $\epsilon{=}0.05$, $C{=}5$,
$h=H(\bar p)/\log C$, $q=|\mathcal B|^{-1}\sum_i\max_c p_{ic}$,
$\phi(h)=(1-h)[1-\tfrac{8}{3}[h-0.7]_+]$, and $[u]_+=\max(u,0)$.
The entropy-diversity term grows whenever a batch's predictions concentrate, and we
attenuate it above normalized entropy 0.7 so it stops pushing once the head is
healthy. Setting $\lambda{=}0$ removes that term alone, while the confidence
penalty $5[q-0.9]_+$ stays fixed everywhere. The class-balanced sampler and the
entropy-diversity term are the two anti-collapse components we ablate in
Section~\ref{sec:results}. Predicted-class diversity is the fraction of the five
labels that ever appear as a validation-set argmax.

Each round, the server broadcasts $\theta^{(t)}$, every eligible client runs three
local AdamW epochs on class-balanced mini-batches under eq.~\eqref{eq:loss}, clips the
gradient norm at one, and returns its trainable tensors for the sample-weighted
update in eq.~\eqref{eq:fedobj}. Operational runs start from public DistilBERT and
ViT-Base/16 weights with random task heads. We fine-tune both encoders and keep
unused unimodal heads and the inactive CNN fallback out of optimization and
communication.

\subsection{Controlled Proxy Corpus}

\textbf{Two clinical image corpora.} Dataset~A is the corpus behind our first run set. Its 3,000
images are procedurally generated X-ray-style patterns rather than radiographs, and
its text labels were read off the passages later fed to the classifier. Dataset~B
replaces every image with a public radiograph and keeps the notes unchanged, at 600
examples per class under the same split. We report the two separately.

In Dataset B, we built a balanced 3,000-image proxy corpus at 600 examples per class. The
radiographs come from the Kermany pneumonia collection~\cite{Kermany2018}, the NIH
ChestX-ray dataset~\cite{Wang2017chestxray8}, and the COVID-19 Radiography
Database~\cite{Chowdhury2020covid}, and our loader logs verify 3,000 public images
across all five classes. Each class is drawn from the collection that labels it,
which leaves the image axis partly confounded with source collection. Images are
resized to
$224{\times}224$, ImageNet-normalized, shuffled deterministically, and split 80/20
into 2,400 training and 600 validation examples.

\textbf{Clinical text data.}
We use no patient electronic health record (EHR) text, and the notes come from a
template engine. It pairs each image label, at
the class level, with a synthetic clinical-style note assembled from observation,
symptom, context, and indicator slots. To keep the network from memorizing an
obvious keyword we engineered the mix deliberately: 25\% of notes
draw every slot from the target class, 35\% mix target and non-target slots, and
40\% are heavily mixed while retaining a probabilistic target-class indicator.
This cuts direct label wording, but image and note do not come from the same
patient and must never be described as a real image--report pair.

\label{sec:corpus}
The corpus supports internal comparison, not absolute claims.
Every configuration (one choice of aggregation rule, fusion operator,
initialization, and anti-collapse components) trains and is scored on the same
corpus, so a gap between two of them is
attributable to the factor under test rather than to the data. Absolute scores carry
no such guarantee. Templates can leave
shortcuts, label-level pairing strips out the discordance and missingness of real
records, each class is drawn from a different source collection, and the split
is grouped by neither patient nor source.

\subsection{Experimental Setup}

\textbf{Hyperparameters.}
Table~\ref{tab:hyper} collects the protocol. Every client trains with AdamW
(learning rate $10^{-4}$, weight decay $0.01$) for three local epochs per round.
The suite runs eight rounds at batch size 16 in FP32, without automatic mixed
precision (AMP). Our reference setting is five clients at Dirichlet $\alpha=1.0$,
and Section~\ref{sec:results} varies $\alpha$ and the client count explicitly. Seeds
0 and 1 drive initialization, sampling, and partitioning, with one exception:
pooled-start reuses a single seed-0 checkpoint under seed-specific FL partitions. We
ran everything in PyTorch on one NVIDIA H100 NVL device. The GPU-synchronized timer
wraps sequential local training and
aggregation within a round, and excludes validation, setup, I/O, networking,
concurrency, and multi-node throughput. Our timings are an execution-cost record,
not distributed throughput. We also did not force deterministic CUDA algorithms, so
a seed fixes the software pseudorandom streams without buying identical reruns.

\textbf{Partitioning.}
For each class, a Dirichlet draw allocates training indices across clients. Smaller
$\alpha$ concentrates client label distributions, and $\alpha\!\to\!\infty$
approaches independent and identically distributed (IID)
allocation~. A nominal $\alpha$ does not fully describe a finite split,
so we report the per-client class shares we actually drew alongside our scalability
results. The split is fixed across rounds for a given seed. Shards
holding fewer than four examples skip local optimization but still contribute the
unchanged global state at their sample weight. We do not renormalize over active
clients, which is why our cells report active alongside nominal $K$.

\textbf{Matched recent-method controls.}
Our FedMME-style baseline uploads each client model once and applies the one-shot
voting structure in~\cite{Wang2025fedmme}. Our implementation uses equal-weight hard
voting and breaks ties by mean softmax. We run it at two local
budgets. Our matched 24 epochs isolates the method from the budget; FedMME's native
100 epochs removes the budget discrepancy. Reporting both keeps the two confounds
separate. Either way it keeps the common corpus,
shard profile, encoders, and optimizer, substituting our five-class inputs for
FedMME's generated-report pipeline. The P-FIN-style stress test fixes
three multimodal and two image-only clients, projects both encoders to
256-dimensional normalized features, and compares zero filling, deterministic FIN,
Gaussian $\beta$-negative-log-likelihood ($\beta$-NLL) imputation ($\beta{=}0.5$),
and uncertainty-weighted aggregation with Fed-UQ-Avg balance weight 0.6 and
temperature 0.2~\cite{Shahid2026pfin}. We kept concatenation instead of P-FIN's
bidirectional cross-modal attention: with one pooled feature per modality,
cross-attention carries unit weight and collapses to a linear projection. FedProx
uses $\mu{=}0.01$. Each ablation changes one component and leaves every other
setting at the shared defaults that follow, most of which are fixed by what the
hardware allows. Three local epochs across eight rounds
give a tractable 24-local-epoch budget, and batch size 16 is what holds both encoders
plus one local model copy in GPU memory for the largest FP32 model. The grid crosses $K{=}\{3,5,10,20\}$ with $\alpha{=}\{0.1,1,5\}$ and the
$K{=}5$ sweep adds $\alpha{=}0.3$ and $0.5$, so the study spans severe to near-IID
skew; $K{=}5$ is our small cross-silo reference and $K{=}3$--20 probes count
sensitivity. Two seeds support descriptive claims, not significance. Everything else
in Table~\ref{tab:hyper} is a fixed default, not a per-model optimum.

\begin{table}[t]
\caption{Common simulation parameters.}
\label{tab:hyper}
\centering\footnotesize
\setlength{\tabcolsep}{4pt}
\begin{tabular}{llll}
\toprule
\textbf{Parameter} & \textbf{Value} & \textbf{Parameter} & \textbf{Value} \\
\midrule
Local Epochs ($E$)   & 3      & Seeds                    & 0, 1 \\
Fed.\ Rounds ($T$)   & 8      & Grid $K$ (ref.)          & 3, 5, 10, 20 (5) \\
Batch Size           & 16     & Grid $\alpha$ (ref.)     & 0.1, 1, 5 (1) \\
AdamW LR             & $10^{-4}$ & $K{=}5$ extra $\alpha$ & 0.3, 0.5 \\
Weight Decay         & $0.01$ & Precision                & FP32, no AMP \\
Diversity wt.\ ($\lambda$) & 1.0 & Init.                & Public enc.\ + rand.\ heads \\
\bottomrule
\end{tabular}
\end{table}

\section{Results}
\label{sec:results}

\subsection{Matched Baselines under Severe Skew}

\begin{figure}[!t]
\centering
\begin{minipage}[t]{0.425\columnwidth}\centering
\includegraphics[width=\linewidth]{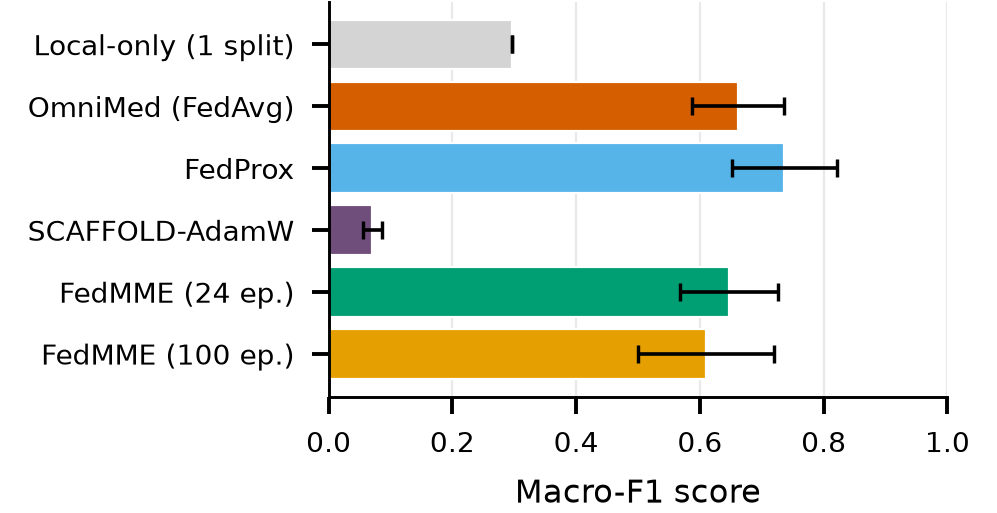}\\[-5pt]
{\scriptsize\textbf{(a)}}
\end{minipage}\hfill
\begin{minipage}[t]{0.425\columnwidth}\centering
\includegraphics[width=\linewidth]{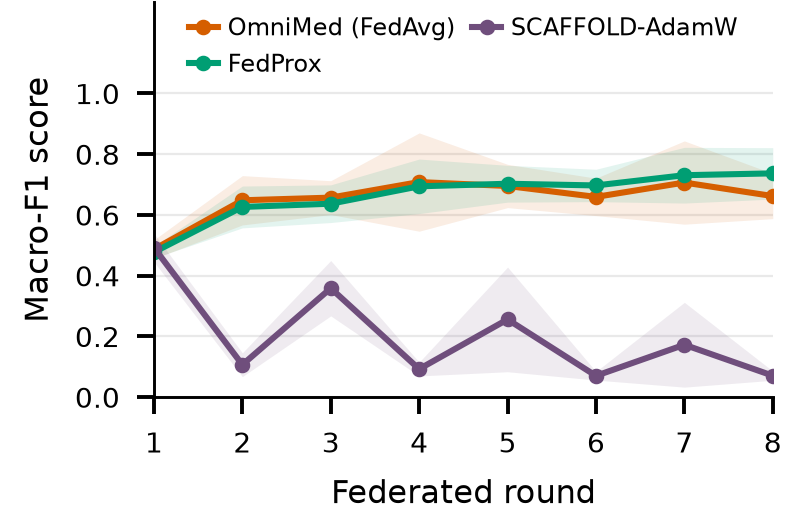}\\[-5pt]
{\scriptsize\textbf{(b)}}
\end{minipage}\\[1pt]
\begin{minipage}[t]{0.425\columnwidth}\centering
\includegraphics[width=\linewidth]{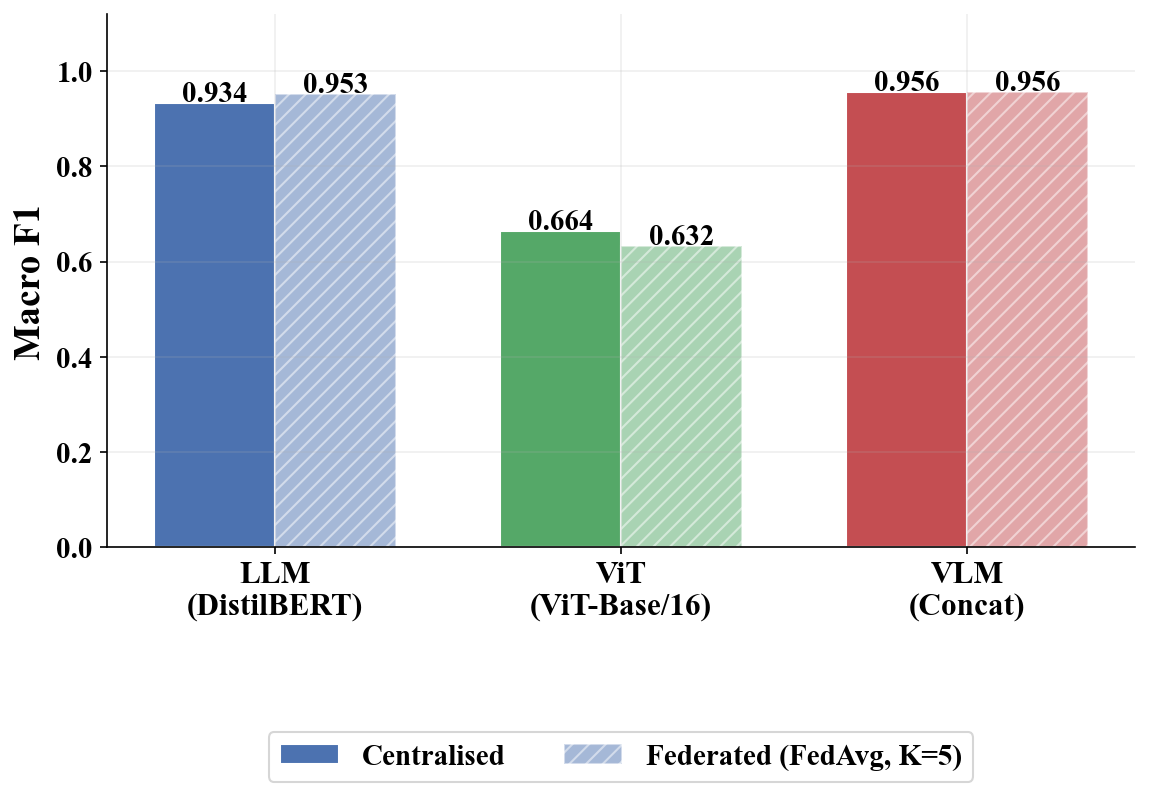}\\[-5pt]
{\scriptsize\textbf{(c)}}
\end{minipage}\hfill
\begin{minipage}[t]{0.425\columnwidth}\centering
\includegraphics[width=\linewidth,trim=0 0 875bp 0,clip]{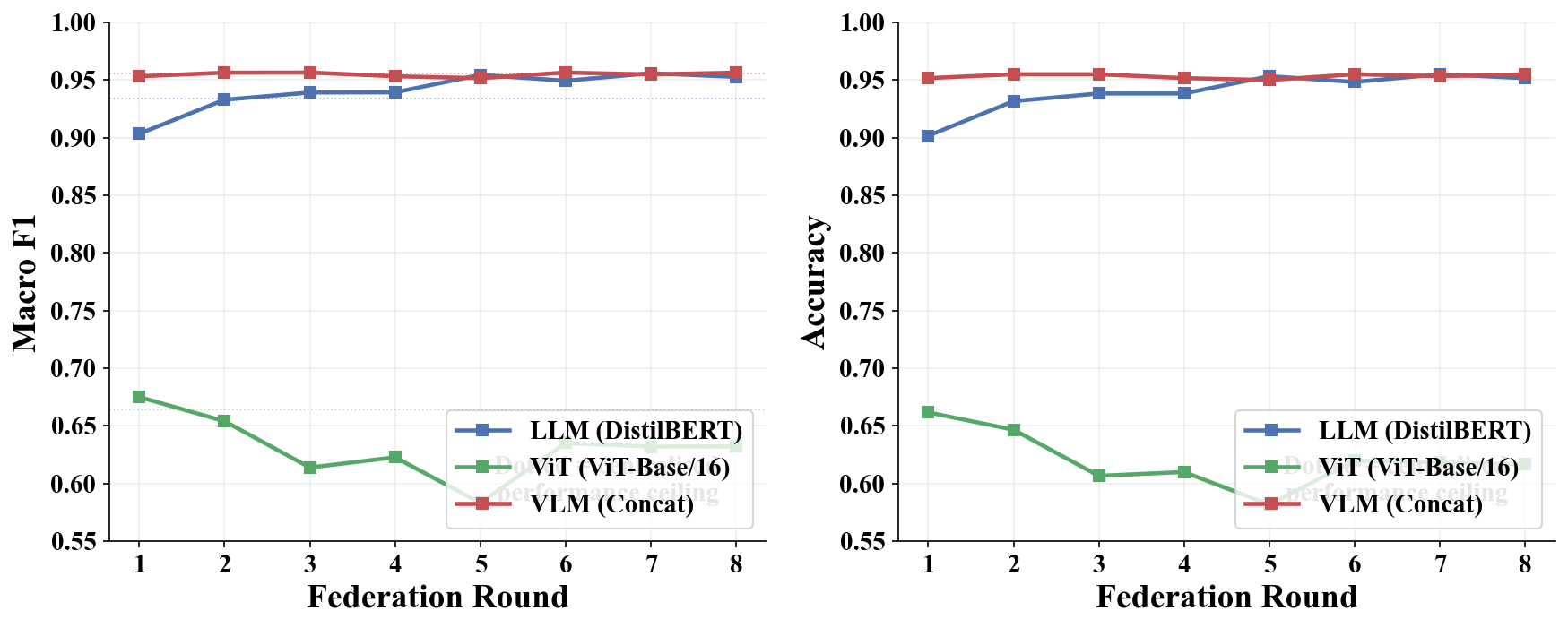}\\[-5pt]
{\scriptsize\textbf{(d)}}
\end{minipage}
\caption{(a) Matched baselines on Dataset~B at $\alpha{=}0.1$, $K{=}5$.
(b) Convergence under severe skew. (c) Centralized against FedAvg on Dataset~A.
(d) Federated convergence on Dataset~A.}
\label{fig:baseline}
\end{figure}

Every result below is on Dataset~B unless it names Dataset~A.
Every configuration in Fig.~\ref{fig:baseline}(a) shares the corpus, partition draw,
encoders,
and optimizer at $\alpha{=}0.1$, $K{=}5$; all but the native-budget FedMME run also
share the 24-local-epoch budget. Local-only training averages a macro-F1 score of
0.297 across the five clients of one partition. Over
two matched seeds, FedAvg reaches $0.662\!\pm\!0.074$ and FedProx
$0.737\!\pm\!0.085$. Their 0.075 gap sits inside the wider of the two sample
standard deviations, so these runs do not separate the two methods. Our
SCAFFOLD--AdamW adaptation lands at $0.070\!\pm\!0.015$, and the
oscillation in Fig.~\ref{fig:baseline}(b) points at our AdamW control-variate adaptation rather than
at classical SGD-based SCAFFOLD. The FedMME-style baseline trains five client models
independently and combines them by equal-weight hard voting with mean-softmax ties. At our matched 24-epoch budget it reaches
$0.647\!\pm\!0.080$ (seeds 0.703, 0.591), and at FedMME's native 100-epoch budget
it reaches $0.609\!\pm\!0.109$ (seeds 0.687, 0.532). Quadrupling local computation
does not close the gap to iterative aggregation, so the gap belongs to one-shot
aggregation under severe skew, not to a budget handicap. One aggregation has no later round in which to repair a
poor local solution, and both budgets stay below FedAvg and FedProx. Every configuration except
SCAFFOLD--AdamW clears
local-only training, though the overlapping intervals make that a coarse ordering
rather than a ranking.

\textit{Dataset~A.} On the synthetic-image corpus, centralized
text/image/concatenation F1-score was 0.934/0.664/0.956 against FedAvg
0.953/0.632/0.956, and multimodal fusion led both unimodal branches. Neither its
text score nor its image score is an independent measurement of a radiograph
task, so we draw no Dataset~B conclusion from it.

\subsection{Anti-Collapse, Fusion, and Initialization Ablations}

\begin{figure}[!t]
\centering
\includegraphics[width=0.98\columnwidth,height=0.80in,keepaspectratio]{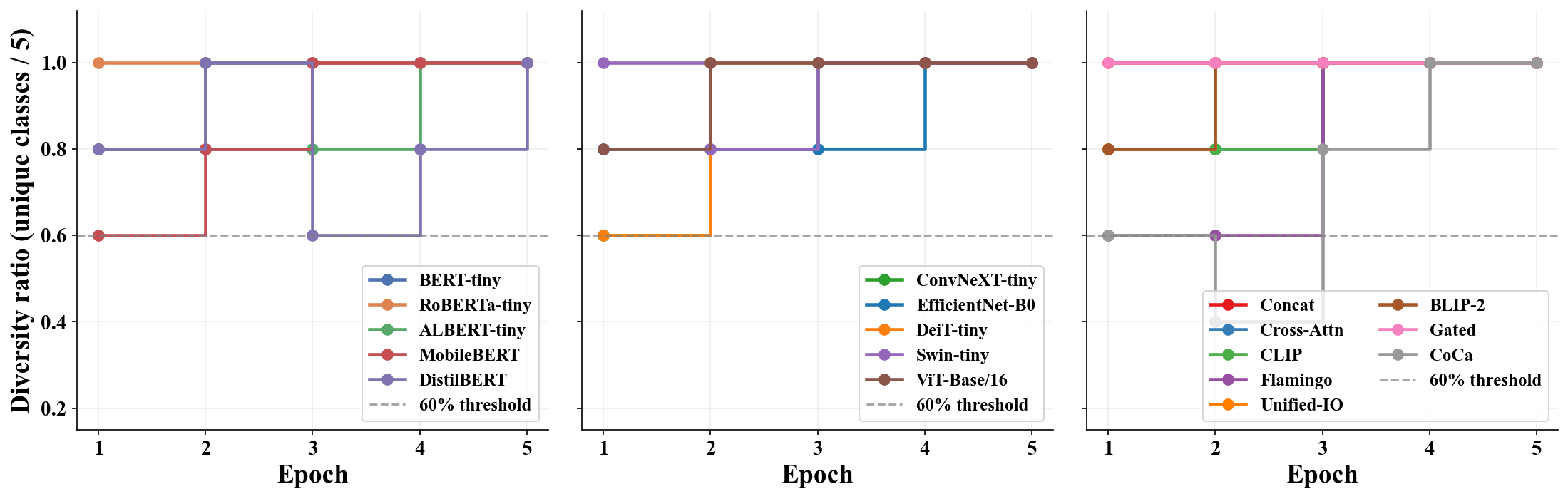}
\caption{Per-epoch class diversity, the fraction of the five classes predicted, on
Dataset~A.}
\label{fig:diversity}
\end{figure}

\begin{figure}[!t]
\centering
\begin{minipage}[t]{0.415\columnwidth}\centering
\includegraphics[width=\linewidth]{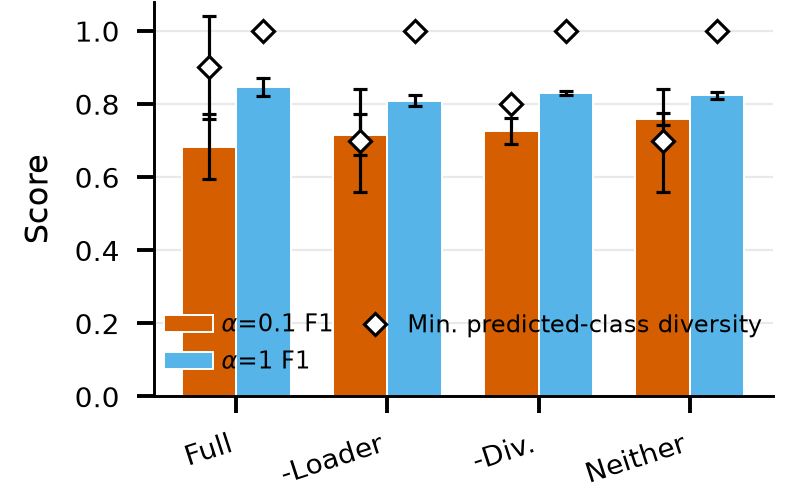}\\[-5pt]
{\scriptsize\textbf{(a)}}
\end{minipage}%
\hfill
\begin{minipage}[t]{0.475\columnwidth}\centering
\includegraphics[width=\linewidth]{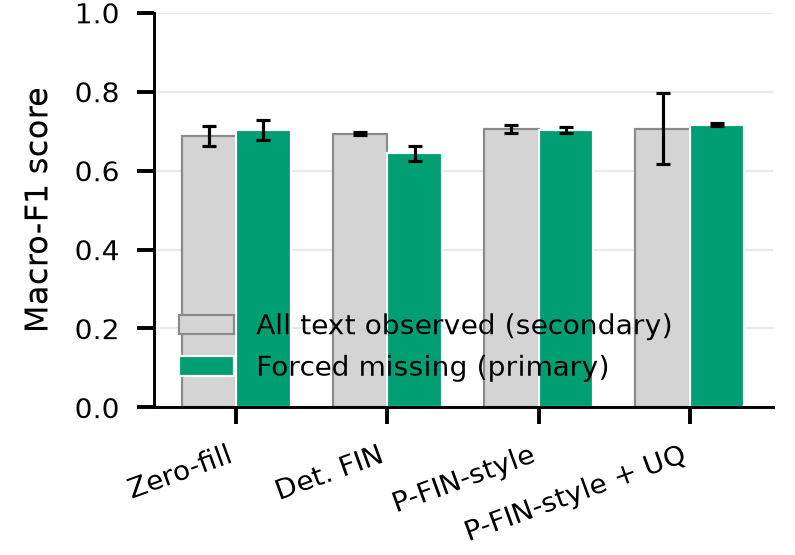}\\[-5pt]
{\scriptsize\textbf{(b)}}
\end{minipage}
\caption{(a) Anti-collapse F1 and diversity under severe skew.
(b) P-FIN-style forced-missing-text results.}
\label{fig:anti_missing}
\end{figure}

Under severe skew the full class-balanced-sampler and entropy-diversity stack
reaches $0.683\!\pm\!0.089$ F1 at minimum predicted-class diversity 0.90. Drop the
sampler, the entropy-diversity term, or both, and F1 rises to
$0.716\!\pm\!0.055$, $0.726\!\pm\!0.035$, and $0.758\!\pm\!0.016$, while minimum
diversity falls to 0.70, 0.80, and 0.70
  (Fig.~\ref{fig:anti_missing}(a)). The stack buys class coverage and costs accuracy, and
its 0.075 F1 deficit to the unregularized configuration sits inside its own seed
spread. The
confidence penalty in \eqref{eq:loss} stays fixed
across all four. At $\alpha{=}1$ the same four configurations land within
$0.809$--$0.846$ with every minimum diversity at 1.0. On real radiographs these
components no longer earn their place on F1. They remain a diversity guard under
severe skew and are inert at moderate skew. One more
observation: three suites nominally repeat the same severe-skew
FedAvg/concatenation setting, yet return $0.643\!\pm\!0.006$, $0.662\!\pm\!0.074$,
and $0.683\!\pm\!0.089$, a 0.040 span that the widest of those spreads covers. We
report them separately rather than pool them or quote the flattering one.
Figure~\ref{fig:diversity} shows the per-epoch class diversity on Dataset~A.

\begin{figure}[!t]
\centering
\begin{minipage}[t]{0.425\columnwidth}\centering
\includegraphics[width=\linewidth]{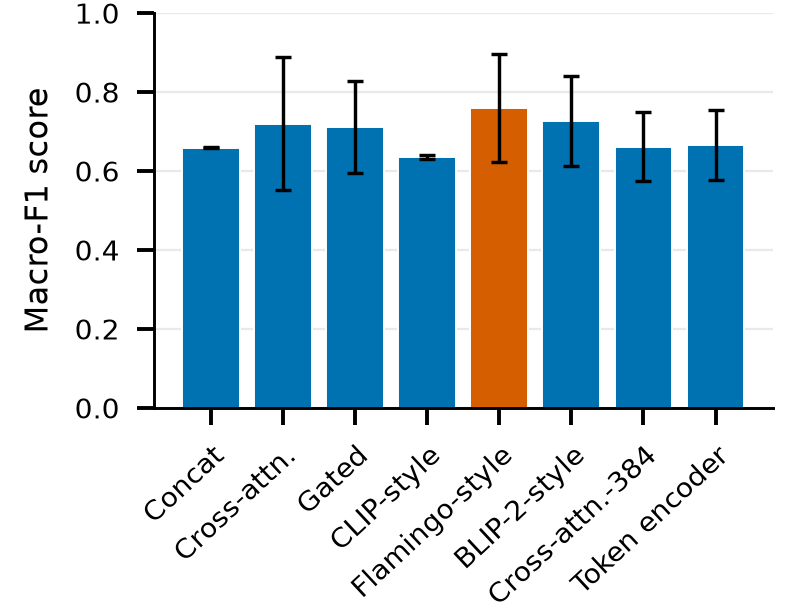}\\[-5pt]
{\scriptsize\textbf{(a)}}
\end{minipage}\hfill
\begin{minipage}[t]{0.425\columnwidth}\centering
\includegraphics[width=\linewidth]{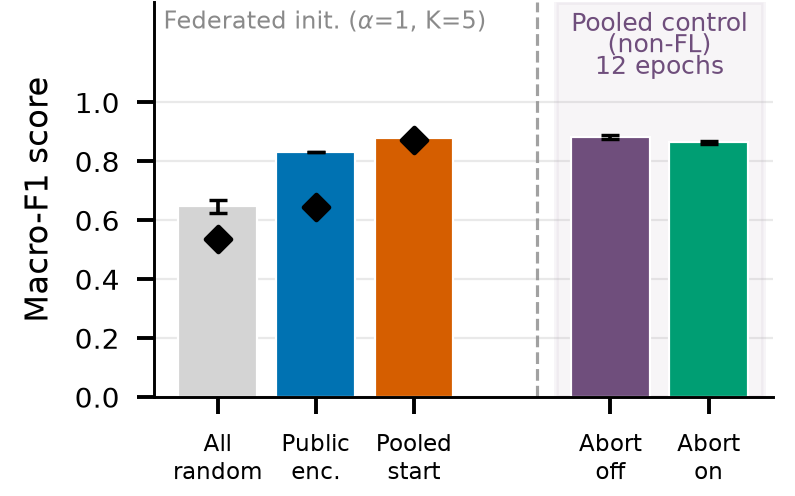}\\[-5pt]
{\scriptsize\textbf{(b)}}
\end{minipage}\\[1pt]
\begin{minipage}[t]{0.425\columnwidth}\centering
\includegraphics[width=\linewidth]{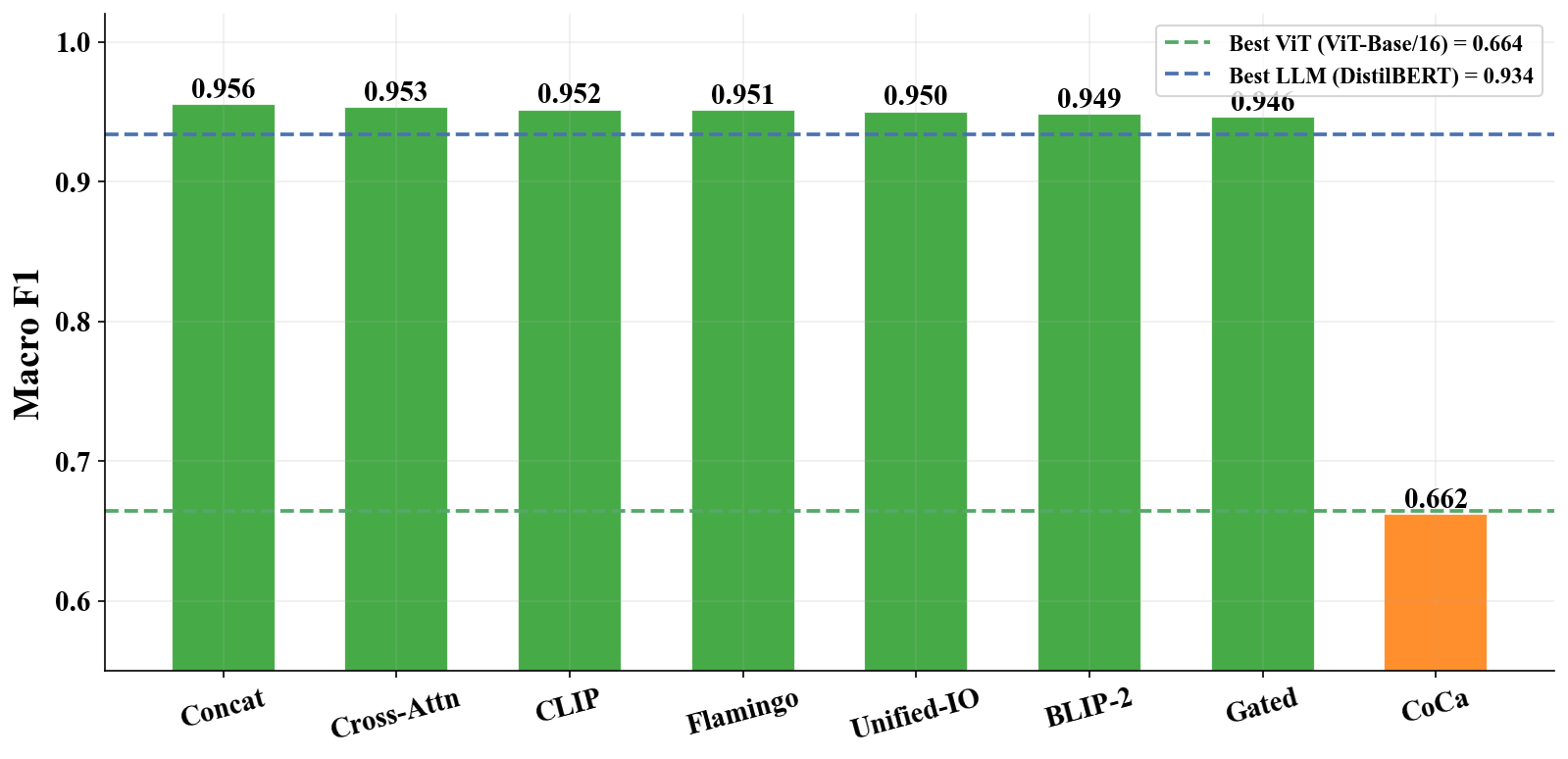}\\[-5pt]
{\scriptsize\textbf{(c)}}
\end{minipage}\hfill
\begin{minipage}[t]{0.425\columnwidth}\centering
\includegraphics[width=\linewidth]{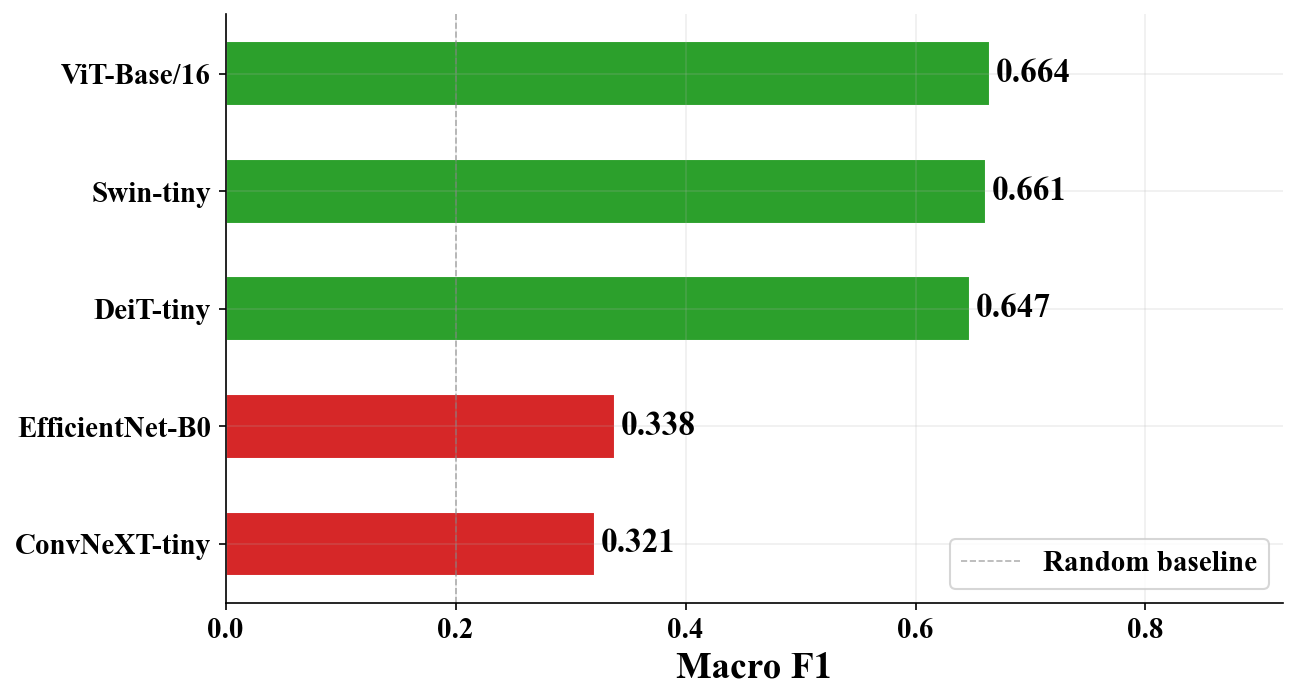}\\[-5pt]
{\scriptsize\textbf{(d)}}
\end{minipage}
\caption{(a) Fusion rules and (b) initialization controls on Dataset~B.
(c) Fusion sweep and (d) vision-backbone sweep on Dataset~A.}
\label{fig:fusion}
\end{figure}

Figure~\ref{fig:fusion}(a) trains all eight fusion rules
inside the operational federated loop. The
means run from $0.636$ for CLIP-style fusion up to
$0.759$ for the Flamingo-style rule, with per-rule sample standard
deviations of $0.000$--$0.167$. Since the widest interval exceeds the entire
between-rule spread, the sweep cannot rank fusion operators, and we will not. Projected concatenation, our default everywhere else, sits mid-pack at
$0.659\!\pm\!0.000$.

Figure~\ref{fig:fusion}(b) separates three federated
initializations at $\alpha{=}1$. The all-random configuration
keeps both architectures but initializes every weight from scratch, opening at 0.535
round-1 F1 and closing at $0.647\!\pm\!0.022$. Public
encoders with random task modules, our operational default, open at 0.646 and close
at $0.831\!\pm\!0.001$. Pooled-start reuses one seed-0 checkpoint from a 12-epoch
centralized run for both FL seeds, so its pretraining sits outside the shared
budget; it opens far higher at 0.874 and ends higher at $0.877\!\pm\!0.007$. Public encoders buy roughly 0.18 F1 over random
initialization, and the pooled checkpoint, selected on the reporting split and so a
diagnostic rather than an unbiased estimate, keeps its head start through round~8.
Non-federated abort-off/on controls post $0.899\!\pm\!0.022$ and
$0.871\!\pm\!0.012$ at diversity 1.0, so the branch never fires and the gap is rerun
variation.

Figure~\ref{fig:anti_missing}(b) is our matched P-FIN-style adaptation, where 256-dimensional
$L_2$-normalized features and concatenation plus an MLP replace the source
bidirectional-attention stack. On the primary forced-missing-text metric the four
imputation rules land within 0.018 of each other: zero filling $0.688\!\pm\!0.025$,
deterministic FIN $0.693\!\pm\!0.004$, Gaussian $\beta$-NLL imputation
$0.706\!\pm\!0.010$, and uncertainty-weighted aggregation $0.706\!\pm\!0.089$. The
means order as P-FIN reports, but the uncertainty-weighted rule ties the plain
probabilistic one and is the noisiest of the four, swinging from 0.770 to 0.643
across seeds, so these runs do not establish that modeling uncertainty tightens the
spread.

\subsection{Retrieval-Stage Evaluation for Retrieval-Augmented Generation (RAG)}

\begin{figure}[!t]
\centering
\begin{minipage}[t]{0.475\columnwidth}\centering
\includegraphics[width=\linewidth]{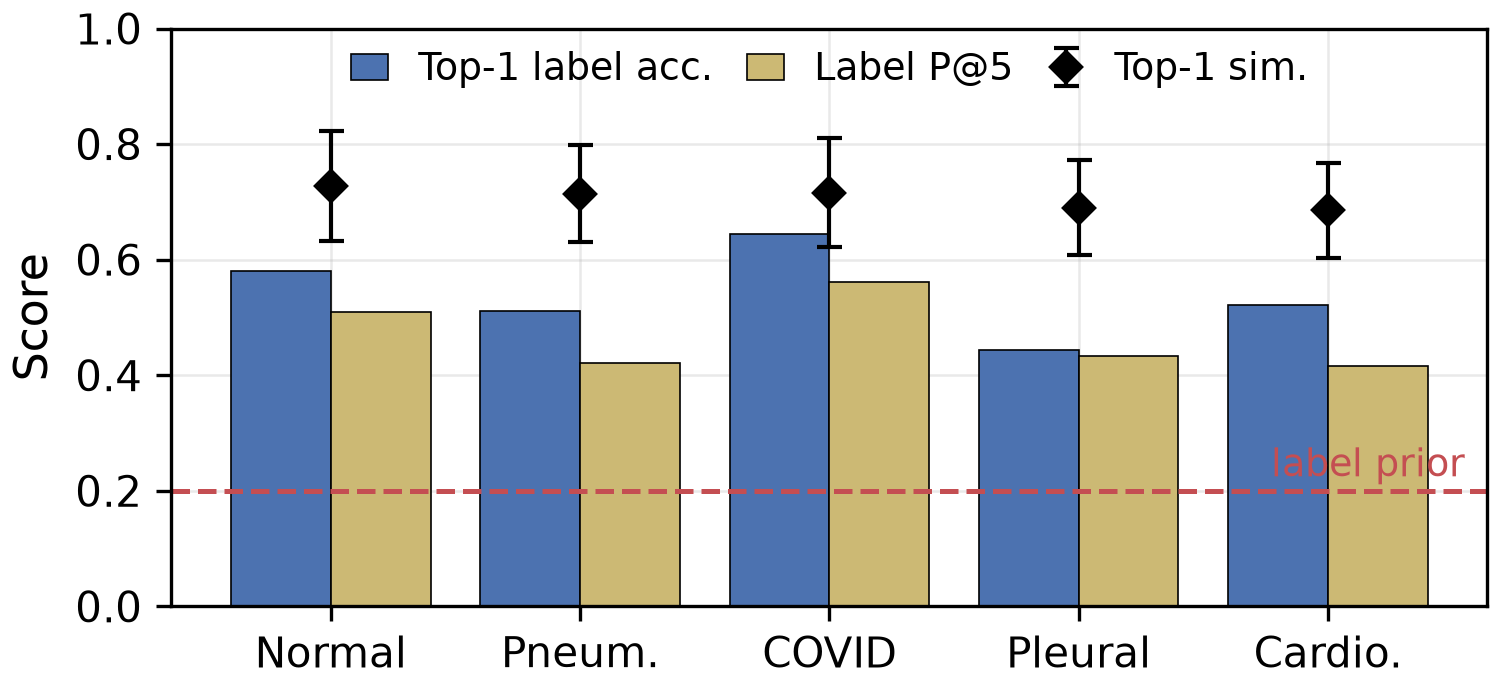}\\[-5pt]
{\scriptsize\textbf{(a)}}
\end{minipage}\hfill
\begin{minipage}[t]{0.38\columnwidth}\centering
\includegraphics[width=\linewidth]{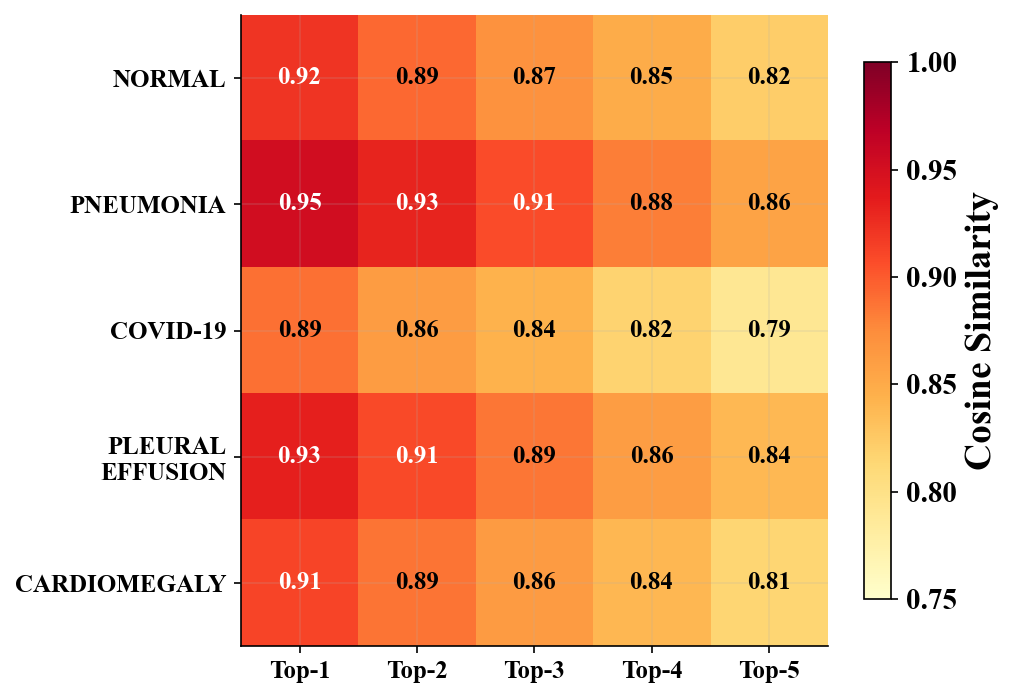}\\[-5pt]
{\scriptsize\textbf{(b)}}
\end{minipage}
\caption{(a) Retrieval metrics by condition on Dataset~B.
(b) Top-$k$ class-similarity matrix on Dataset~A.}
\label{fig:retrieval}
\end{figure}

Exact inner-product FAISS search~\cite{Johnson2021faiss} over training-fitted
TF--IDF note vectors gives condition-macro averages of 0.540 for top-1 same-label
retrieval accuracy, 0.468 for same-label precision@5, and 0.707 for mean top-1
cosine similarity, with the five condition rows covering all 600 validation
queries (Fig.~\ref{fig:retrieval}(a)). Top-1 accuracy sits well above the 0.2 label prior yet
far below the 0.707 mean similarity, the signature of a template corpus whose
neighbors are lexically close whether or not they share a label. This check
establishes no clinical retrieval quality.

\subsection{Scalability, Communication, and Resource Cost}

\begin{figure}[!t]
\centering
\begin{minipage}[t]{0.425\columnwidth}\centering
\includegraphics[width=\linewidth]{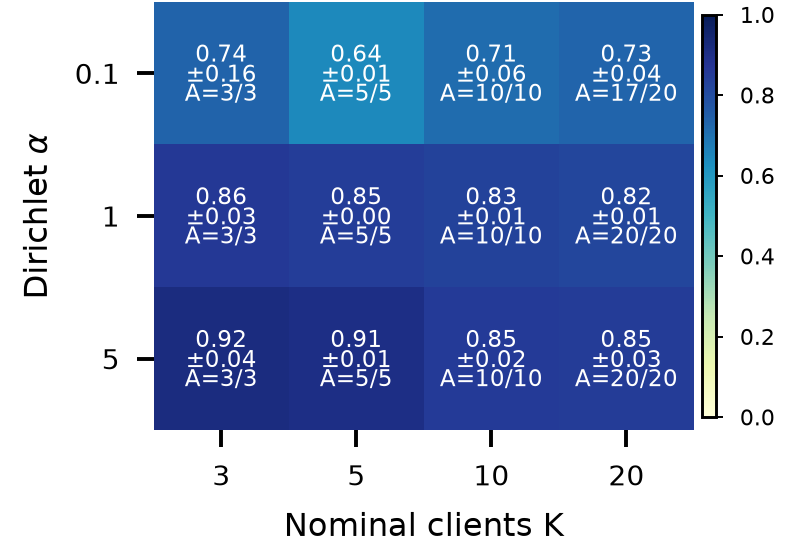}\\[-5pt]
{\scriptsize\textbf{(a)}}
\end{minipage}\hfill
\begin{minipage}[t]{0.425\columnwidth}\centering
\includegraphics[width=\linewidth]{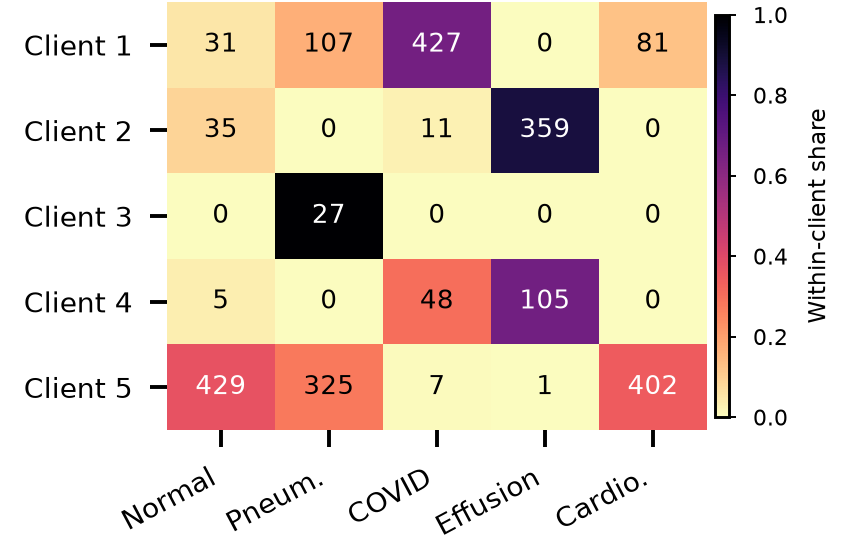}\\[-5pt]
{\scriptsize\textbf{(b)}}
\end{minipage}
\caption{(a) Macro-F1 over client count and skew. (b) One severe-skew
client allocation.}
\label{fig:scale_accuracy}
\end{figure}

Figure~\ref{fig:scale_accuracy}(a) assembles the full Cartesian product of
$K{=}\{3,5,10,20\}$ and $\alpha{=}\{0.1,1,5\}$, and one factor dominates
throughout: label skew, not client count. Every $\alpha{=}5$ cell lies in
$0.849$--$0.916$ and every $\alpha{=}1$ cell in $0.818$--$0.862$, whereas the
$\alpha{=}0.1$ row drops to $0.643$--$0.738$ with far wider spreads, up to
$\pm0.159$ at $K{=}3$. Within a row, moving $K$ by a factor of nearly seven shifts
the mean by at most 0.10 F1, and the $\alpha{=}0.1$ row is not even monotone in
$K$. Our auxiliary $K{=}5$ checks at $\alpha{=}0.3$ and $0.5$ give
$0.814\!\pm\!0.001$ and $0.796\!\pm\!0.037$. Cells report $A/K$, where $A$ counts
clients whose realized shard reaches four samples: allocation is complete except
under severe skew, where $K{=}20$ retains 17 active
clients. Figure~\ref{fig:scale_accuracy}(b) shows the realized per-client class counts behind one
severe-skew cell, where single clients hold entire classes.

\begin{figure}[!b]
\centering
\begin{minipage}[t]{0.425\columnwidth}\centering
\includegraphics[width=\linewidth]{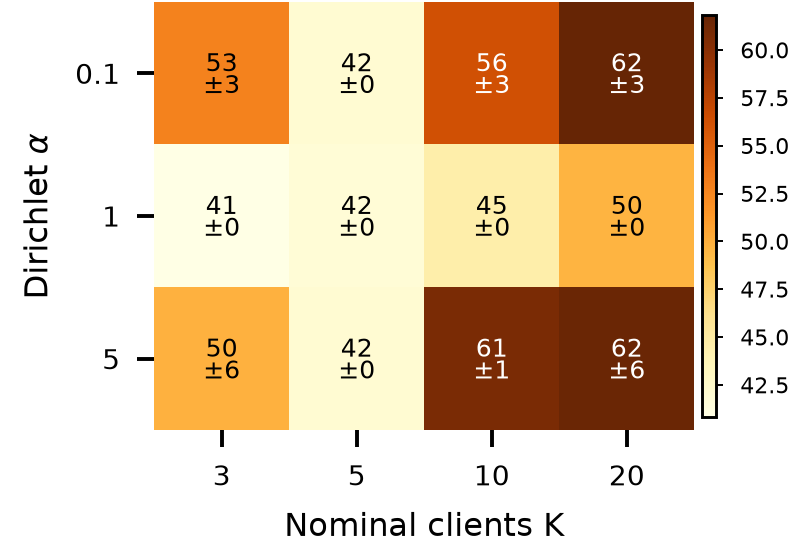}\\[-5pt]
{\scriptsize\textbf{(a)}}
\end{minipage}\hfill
\begin{minipage}[t]{0.425\columnwidth}\centering
\includegraphics[width=\linewidth]{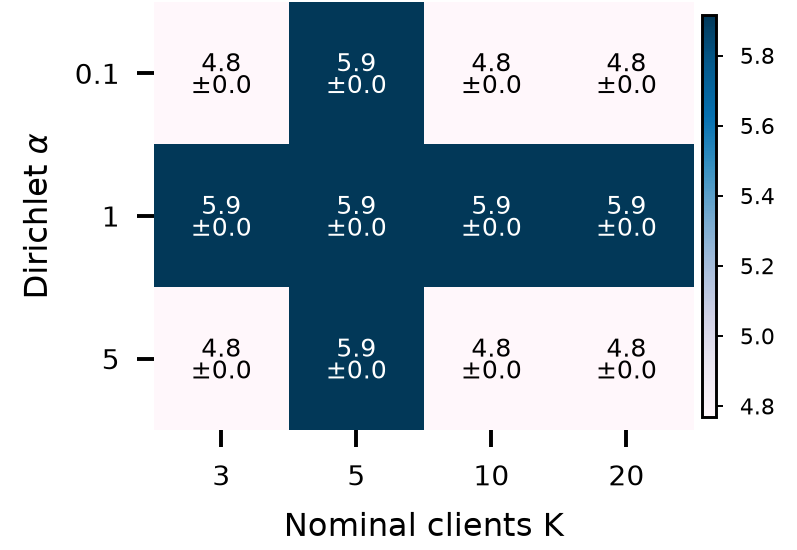}\\[-5pt]
{\scriptsize\textbf{(b)}}
\end{minipage}
\caption{(a) GPU-synchronized round time. (b) Peak allocated memory.}
\label{fig:resources}
\end{figure}

Figure~\ref{fig:resources} reports runtime (left) and peak allocated memory (right)
for the timed block on
one shared H100 NVL. Timed sections span 41--66\,s per round and peak memory
4.77--5.92\,GiB. Neither panel is a scaling curve. Sequential
simulation and contention on a shared server, not client count, explain the
spread.

\begin{figure}[!t]
\centering
\begin{minipage}[t]{0.425\columnwidth}\centering
\includegraphics[width=\linewidth]{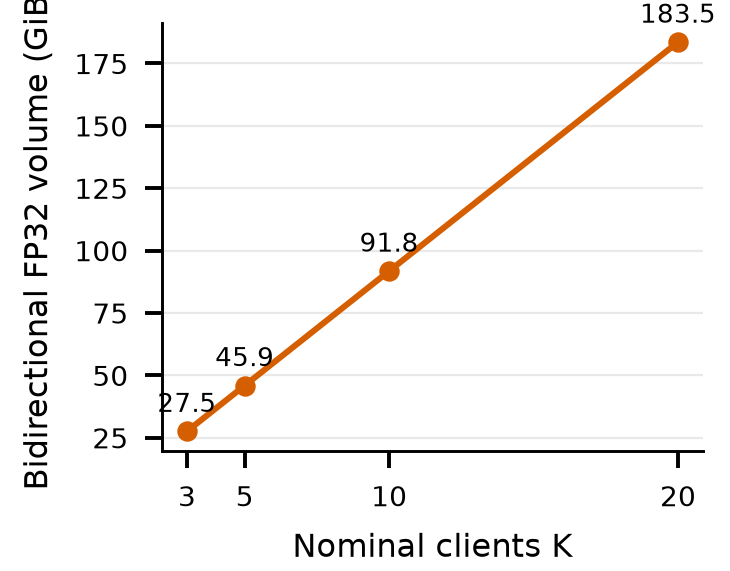}\\[-5pt]
{\scriptsize\textbf{(a)}}
\end{minipage}\hfill
\begin{minipage}[t]{0.425\columnwidth}\centering
\includegraphics[width=\linewidth]{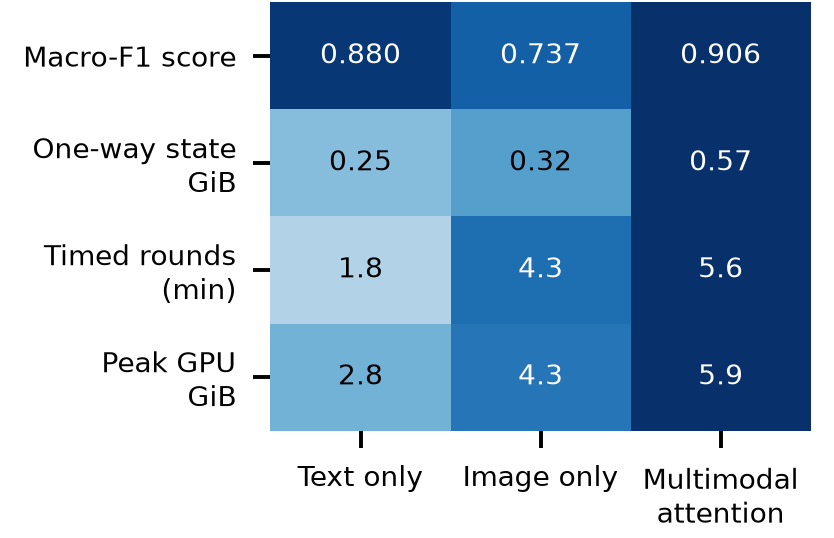}\\[-5pt]
{\scriptsize\textbf{(b)}}
\end{minipage}
\caption{(a) Formula-derived communication volume. (b) Modality-branch
cost.}
\label{fig:costs}
\end{figure}

For $|\theta|$ trainable parameters, $b$ bytes/parameter, $T$ rounds, and
nominal $K$, the plotted bidirectional volume is
\begin{equation}
V_{\mathrm{nom}}=2KT|\theta|b,
\label{eq:comm}
\end{equation}
Here $|\theta|{=}153{,}935{,}621$, $b{=}4$, and $T{=}8$, so a 615,742,484-byte
model state gives nominal bidirectional volumes of 27.5/45.9/91.8/183.5\,GiB at
$K{=}3/5/10/20$ ($1$ GiB$=2^{30}$ bytes). We keep the nominal count when $A{<}K$.
Serialization, secure aggregation~\cite{Ergun2022globecom}, compression, and
algorithm-specific state sit
outside it. Figure~\ref{fig:costs}(a) puts the
operating point plainly. Communication grows linearly in $K$ while accuracy stays
nearly flat in $K$. At $\alpha{=}5$, going from $K{=}3$ to $K{=}20$ costs 0.067 F1
for $6.7\times$ the traffic.

The branch comparison in Fig.~\ref{fig:costs}(b) quantifies the multimodal
overhead on Dataset~B. Text, image, and multimodal F1 are 0.880/0.737/0.906, the
multimodal figure using the residual cross-attention operator our pooled screen
selects; projected concatenation, whose costs the resource rows report, scores
0.813. One-way FP32 model-state sizes are 0.248/0.324/0.573\,GiB, timed round
durations 1.8/4.3/5.6\,min, and peak memory 2.84/4.33/5.92\,GiB. Multimodal costs
$2.3\times$ the model state, $3.1\times$ the wall time and $2.1\times$ the memory of
the text branch, while scoring 0.026 F1 above text and 0.169 above image. Multimodal is the
highest-scoring branch on Dataset~B, at single seed and $\alpha{=}1$.

\section{Discussion and Limitations}

Partition sensitivity is the most consistent finding of the study. At
$\alpha{=}0.1$ even identically configured FedAvg reruns vary substantially, with
honest updates spreading by $\pm0.159$ F1. That spread also limits robust
aggregation. Poisoning-resistant aggregation methods are designed to down-weight
anomalous updates~\cite{Barros2022globecom}, but under severe label skew such a rule
may also discard genuine minority-class signal. Skew and robustness to
malicious clients therefore pull against each other, and measuring where the balance
falls would require an attacker model that we did not define here.

Several bounds apply to the rest of the results. The client-count study is a
sequential simulation, so it omits concurrency, latency, stragglers, dropout, and
heterogeneous accelerators. Eq.~\eqref{eq:comm} and Fig.~\ref{fig:costs}(b) report
analytic costs rather than end-to-end traffic. Clinical claims stay bounded by
Section~\ref{sec:corpus}, since our F1 and retrieval scores do not estimate
diagnostic safety, generalization, or deployment readiness, and two seeds make our
spreads descriptive. We leave larger multi-site data, more seeds, native-method
replications, and privacy testing to future work.

\section{Conclusion}

We presented OmniMed-FL, a multimodal federated learning benchmark covering 18
model variants for five-class clinical condition classification. At
$\alpha{=}1$ and $K{=}5$, DistilBERT leads the text encoders at $F1{=}0.934$,
ViT-Base/16 leads the vision encoders at $F1{=}0.664$, and concatenation achieves
the best overall score of $F1{=}0.956$, improving by 0.022 over text and 0.292 over
vision. The anti-collapse stack maintains five-class prediction diversity for the
text and multimodal variants, while federated training retains 99.1\% of
centralized performance; Fed-VLM matches its centralized counterpart,, and Fed-LLM
slightly exceeds it. These findings show that warm-start initialization and
anti-collapse regularization preserve strong multimodal performance in the
evaluated setting. The scalability analysis also shows that label skew matters
more than client count, communication grows linearly with participating clients,
and multimodal fusion incurs additional model-state, runtime, and memory costs.

Future work will use patient-paired multi-institutional radiology data, formal
differential privacy auditing, asynchronous aggregation, and end-to-end
communication measurements. OmniMed-FL provides a practical blueprint for
decentralized multimodal learning, but external clinical validation and safeguards
remain necessary before deployment in telehealth or diagnostic workflows.

\FloatBarrier
\bibliographystyle{IEEEtran}
\let\oldbibliography\thebibliography
\renewcommand{\thebibliography}[1]{%
  \oldbibliography{#1}%
  \fontsize{5.9}{6.1}\selectfont
  \setlength{\itemsep}{0.25pt}%
  \setlength{\parskip}{0pt}%
  \setlength{\parsep}{0pt}%
  \setlength{\topsep}{0pt}%
  \setlength{\labelsep}{3pt}%
  \linespread{0.98}\selectfont
}

\section*{Acknowledgment}
The work reported in this paper was carried out, in part, during the tenure of
the INAE Chair Professorship held by the third author. The authors gratefully
acknowledge partial support from the Ministry of Electronics and Information
Technology (MeitY), Government of India, F. No. 13(1)/2024-CC\&BT, and the
Department of Science and Technology (DST), Government of India, under Sanction
Letter Nos. DST/INT/India-EU/RAIDO/2024 (G) and DST/INT/India-EU/RAIDO/2024 (C).

\balance
  \end{document}